\documentclass[10pt,twocolumn,letterpaper]{article}

\usepackage{cvpr}
\usepackage{times}
\usepackage{epsfig}
\usepackage{graphicx}
\usepackage{amsmath}
\usepackage{amssymb}
\usepackage[algoruled,boxed,lined]{algorithm2e}
\usepackage{booktabs}

\usepackage[utf8]{inputenc} 
\usepackage{url}            
\usepackage{booktabs}       
\usepackage{amsfonts}       
\usepackage{nicefrac}       
\usepackage{microtype}      
\usepackage{color}
\usepackage{subcaption}
\graphicspath{ {./images/} }
\usepackage{multirow}

\DeclareMathOperator*{\argmin}{arg\,min}

\usepackage[pagebackref=true,breaklinks=true,letterpaper=true,colorlinks,bookmarks=false]{hyperref}

\cvprfinalcopy 

\def\cvprPaperID{1485} 

\ifcvprfinal\fi
\begin{document}


\title{PADS: Policy-Adapted Sampling for Visual Similarity Learning}


\author{
Karsten Roth\thanks{Authors contributed equally to this work.} \quad \quad Timo Milbich$^*$ \quad \quad Björn Ommer\\
Heidelberg Collaboratory for Image Processing / IWR \\
Heidelberg University, Germany\\
}

\maketitle


\begin{abstract}
Learning visual similarity requires to learn relations, typically between triplets of images. Albeit triplet approaches being powerful, their computational complexity mostly limits training to only a subset of all possible training triplets. Thus, sampling strategies that decide when to use which training sample during learning are crucial. Currently, the prominent paradigm are fixed or curriculum sampling strategies that are predefined before training starts. However, the problem truly calls for a sampling process that adjusts based on the actual state of the similarity representation during training. We, therefore, employ reinforcement learning and have a teacher network adjust the sampling distribution based on the current state of the learner network, which represents visual similarity. Experiments on benchmark datasets using standard triplet-based losses show that our adaptive sampling strategy significantly outperforms fixed sampling strategies. Moreover, although our adaptive sampling is only applied on top of basic triplet-learning frameworks, we reach competitive results to state-of-the-art approaches that employ diverse additional learning signals or strong ensemble architectures. Code can be found under \url{https://github.com/Confusezius/CVPR2020_PADS}.
\end{abstract}

\begin{figure}[t]
\begin{center}
\includegraphics[width=0.43\textwidth]{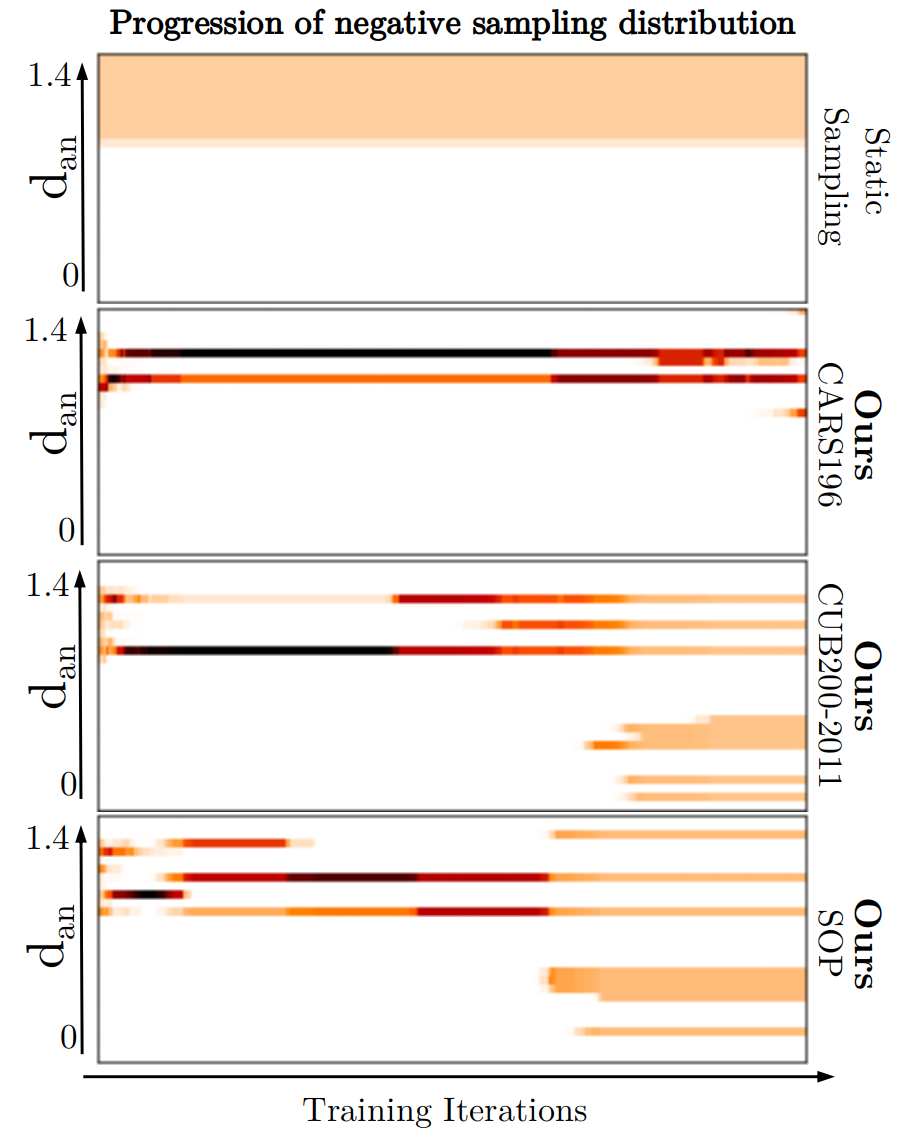}
\end{center}
\caption{\textit{Progression of negative sampling distributions over training iterations.} A static sampling strategy\cite{margin} follows a fixed probability distribution over distances $d_{an}$ between anchor and negative images. In contrast, our learned, discretized sampling distributions change while adapting to the training state of the DML model.
This leads to improvements on all datasets close to $4\%$ compared to static strategies (cf. Tab. \ref{tab:baselines}). Moreover, the progression of the adaptive distributions varies between datasets and, thus, is difficult to model manually which highlights the need for a learning based approach.}
\label{fig:first_page}
\end{figure}

\section{Introduction}
Capturing visual similarity between images is the core of virtually every computer vision task, such as image retrieval\cite{margin,npairs,abier,pr20_reliable_relations}, pose understanding~\cite{milbich2017unsupervised,Coskun2018HumanMA,cliquecnn,suemer_iccv17}, face detection\cite{semihard} and style transfer~\cite{kotovenko_styletransfer}. Measuring similarity requires to find a representation which maps similar images close together and dissimilar images far apart. This task is naturally formulated as Deep Metric Learning (DML) in which individual pairs of images are compared\cite{contrastive,npairs,lifted} or contrasted against a third image\cite{semihard,margin,angular} to learn a distance metric that reflects image similarity. Such triplet learning constitutes the basis of powerful learning algorithms\cite{mic,abier,Sanakoyeu_2019_CVPR,dreml}. However, with growing training set size, leveraging every single triplet for learning becomes computationally infeasible,
limiting training to only a subset of all possible triplets. Thus, a careful selection of those triplets which drive learning best, is crucial. 
This raises the question: How to determine \emph{which} triplets to present \emph{when} to our model during training?
\\
As training progresses, more and more triplet relations will be correctly represented by the model. Thus, ever fewer triplets will still provide novel, valuable information.
Conversely, leveraging only triplets which are hard to learn\cite{semihard,daml,htg} but therefore informative, impairs optimization due to high gradient variance\cite{margin}. Consequently, a reasonable mixture of triplets with varying difficulty would provide an informative and stable training signal. 
Now, the question remains, \textit{when} to present which triplet? Sampling from a fixed distribution over difficulties may serve as a simple proxy\cite{margin} and is a typical remedy in representation learning in general\cite{vae,nat}. However, \emph{(i)} choosing a proper distribution is difficult; \emph{(ii)} the abilities and state of our model evolves as training progresses and, thus, a fixed distribution cannot optimally support every stage of training; and \emph{(iii)} triplet sampling should actively contribute to the learning objective rather than being chosen independently. Since a manually predefined sampling distribution does not fulfill these requirements, we need to \textit{learn} and adapt it while training a representation. 
\\
Such online adaptation of the learning algorithm and parameters that control it during training is typically framed as a teacher-student setup and optimized using Reinforcement Learning (RL). When modelling a flexible sampling process (the student), a controller network (the teacher) learns to adjusts the sampling such that the DML model is steadily provided with an optimal training signal. Fig. \ref{fig:first_page} compares progressions of learned sampling distributions adapted to the DML model with a typical fixed sampling distribution\cite{margin}.
\\
\indent
This paper presents how to learn a novel triplet sampling strategy which is able to effectively support the learning process of a DML model at every stage of training. To this end, we model a sampling distribution so it is easily adjustable to yield triplets of arbitrary mixtures of difficulty. To adapt to the training state of the DML model we employ Reinforcement Learning to update the adjustment policy. Directly optimizing the policy so it improves performance on a held-back validation set, adjusts the sampling process to optimally support DML training. Experiments show that our adaptive sampling strategy significantly improves over fixed, manually designed triplet sampling strategies on multiple datasets.
Moreover, we perform diverse analyses and ablations to provide additional insights into our method.

\section{Related Work}
Metric learning has become the leading paradigm for learning distances between images with a broad range of applications, including image retrieval\cite{proxynca,dvml,margin}, image classification \cite{Feng_2019_CVPR,Zhe2018DirectionalSD}, face verification\cite{semihard,face_verfication_inthewild,sphereface} or human pose analysis\cite{milbich2017unsupervised,Coskun2018HumanMA}. Ranking losses formulated on pairs\cite{npairs,contrastive}, triplets\cite{semihard,margin,angular,htl} or even higher order tuples of images\cite{quadtruplet, lifted,rankedlist} emerged as the most widely used basis for DML~\cite{icml20}. As with the advent of CNNs datasets are growing larger, different strategies are developed to cope with the increasing complexity of the learning problem.
\\
\noindent
\textbf{Complexity management in DML:} The main line of research are negative sampling strategies\cite{semihard,margin,smartmining} based on distances between an anchor and a negative image. FaceNet\cite{semihard} leverages only the hard negatives in a mini-batch. Wu \etal\cite{margin} sample negatives uniformly over the whole range of distances to avoid large variances in the gradients while optimization. Harwood \etal\cite{smartmining} restrict and control the search space for triplets using pre-computed sets of nearest neighbors by linearly regressing the training loss. Each of them successfully enable effective DML training. However, these works are based on fixed and manually predefined sampling strategies. In contrast, we learn an adaptive sampling strategy to provide an optimal input stream of triplets conditioned on the training state of our model.
\\
Orthogonal to sampling negatives from the training set is the generation of hard negatives in form of images\cite{daml} or feature vectors\cite{hardness-aware, htg}. Thus, these approaches also resort to hard negatives, while our sampling process yields negatives of any mixture of difficulty depending on the model state.
\\
Finally, proxy based techniques reduce the complexity of the learning problem by learning one\cite{proxynca} or more \cite{softriple} virtual representatives for each class, which are used as negatives. Thus, these approaches approximate the negative distributions, while our sampling adaptively yields individual negative samples.
\\
\noindent
\textbf{Advanced DML:} Based on the standard DML losses many works improve model performance using more advanced techniques. Ensemble methods \cite{abier,dreml,Sanakoyeu_2019_CVPR} learn and combine multiple embedding spaces to capture more information. HORDE\cite{horde} additionally forces feature representations of related images to have matching higher moments. Roth \etal\cite{mic} combines class-discriminative features with features learned from characteristics shared across classes. Similarly, Lin \etal\cite{dvml} proposes to learn the intra-class distributions, next to the inter-class distribution. All these approaches are applied in addition to the standard ranking losses discussed above. In contrast, our work presents a novel triplet sampling strategy and, thus, is complementary to these advanced DML methods.
\\
\noindent
\textbf{Adaptive Learning:}
Curriculum Learning\cite{curriculum_learning} gradually increases the difficulty of the the samples presented to the model. Hacohen \etal\cite{CL_icml} employ 
a batch-based learnable scoring function to provide a batch-curriculum for training, while we learn how to adapt a sampling process to the training state. Graves \etal\cite{cl_tasks} divide the training data into fixed subsets before learning in which order to use them from training. Further, Gopal \etal~\cite{pmlr-v48-gopal16} employs an empirical online importance sampling distribution over inputs based on their gradient magnitudes during training. Similarly, Shreyas \etal~\cite{data_parameters} learn an importance sampling over instances. In contrast, we learn an online policy for selecting triplet negatives, thus instance relations. Meta Learning aims at learning how to learn. It has been successfully applied for various components of a learning process, such as activation functions\cite{Ramachandran2017SearchingFA}, input masking\cite{neural_data_filter}, self-supervision~\cite{buechler_ECCV_2018}, finetuning~\cite{metaFinetune}, loss functions\cite{ala}, optimizer parameters\cite{optimizer_learning} and model architectures\cite{Pham2018EfficientNA,snas}. 
In this work, we learn a sampling distribution to improve triplet-based learning.

\begin{figure}[t]
\begin{center}
\includegraphics[width=0.4\textwidth]{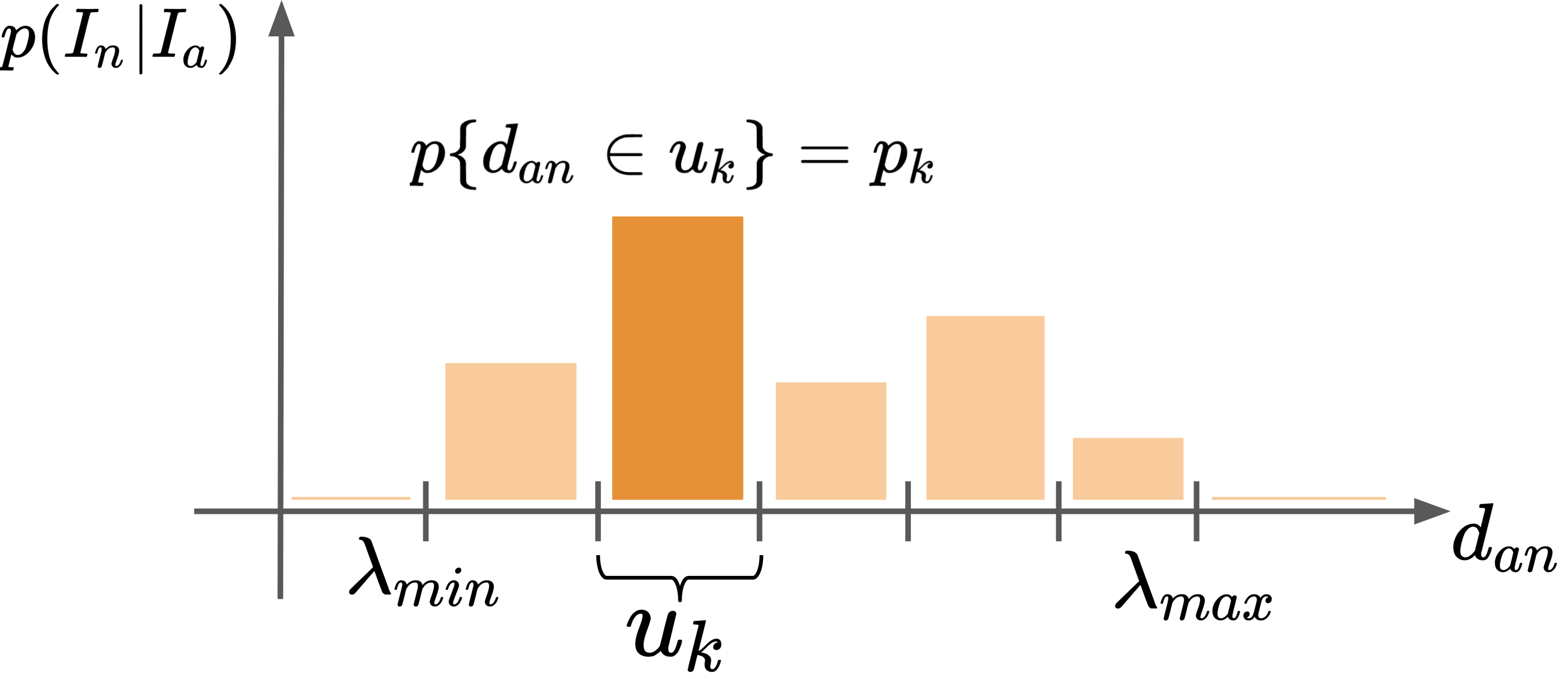}
\end{center}
   \caption{\textit{Sampling distribution $p(I_n|I_a)$}. We discretize the distance interval $U=[\lambda_{\text{min}}, \lambda_{\text{max}}]$ into $K$ equisized bins $u_k$ 
   with individual sampling probabilities $p_k$.}
\label{fig:model_sampling_distr}
\end{figure}

\section{Distance-based Sampling for DML}
Let $\phi_i:= \phi(I_i;\zeta)$ be a $D$-dimensional embedding of an image $I_i \in \mathbb{R}^{H \times W \times 3}$ with $\phi(I_i;\zeta)$ being represented by a deep neural network parametrized by $\zeta$. Further, $\phi$ is normalized to a unit hypersphere $\mathbb{S}$ for regularization purposes \cite{semihard}. 
Thus, the objective of DML is to learn $\phi: \mathbb{R}^{H \times W \times 3} \rightarrow \Phi \subseteq \mathbb{S}$ such that images $I_i, I_j \in \mathcal{I}_{\text{train}}$ are mapped close to another if they are similar and far otherwise, under a standard distance function $d(\phi_i, \phi_j)$. Commonly, $d$ is the euclidean distance, i.e. $d_{ij} := \left\Vert \phi_i -  \phi_j \right\Vert_2$. 
\\
A popular family of training objectives for learning $\phi$ are ranking losses\cite{semihard,margin,npairs,lifted,lifted,contrastive} operating on tuples of images. Their most widely used representative is arguably the triplet loss\cite{semihard} which is defined as an ordering task between images $\{I_a, I_p, I_n\}$, formulated as  

\begin{equation}
\mathcal{L}_{\text{triplet}}(\{I_a, I_p, I_n\};\zeta) = \text{max}(0, d_{ap}^2 - d_{an}^2 + \gamma)
\label{eq:triplet}
\end{equation}

Here, $I_a$ and $I_p$ are the anchor and positive with the same class label. $I_n$ acts as the negative from a different class. Optimizing $\mathcal{L}_{\text{triplet}}$ pushes $I_a$ closer to $I_p$ and further away from $I_n$ as long as a constant distance margin $\gamma$ is violated. 

\subsection{Static Triplet sampling strategies}
\label{sec:static_dist_sampling}
While ranking losses have proven to be powerful, the number of possible tuples grows dramatically with the size of the training set. Thus, training quickly becomes infeasible, turning efficient tuple sampling strategies into a key component for successful learning as discussed here.
\\
When performing DML using ranking losses like Eq.\ref{eq:triplet}, triplets decreasingly violate the triplet margin $\gamma$ as training progresses. Naively employing random triplet sampling entails many of the selected triplets being uninformative, as distances on $\Phi$ are strongly biased towards larger distances $d$ due to its regularization to $\mathbb{S}$. Consequently, recent sampling strategies explicitly leverage triplets which violate the triplet margin and, thus, are difficult and informative. 
\\
\textbf{(Semi-)Hard negative sampling:} Hard negative sampling methods focus on triplets violating the margin $\gamma$ the most, i.e. by sampling negatives $I_n^* = \argmin_{I_n \in \mathcal{I}: d_{an} < d_{ap}} d_{an}$. While it speeds up convergence, it may result in collapsed models\cite{semihard} due to a strong focus on few data outliers and very hard negatives.
Facenet\cite{semihard} proposes a relaxed, semi-hard negative sampling strategy restricting the sampling set to a single mini-batch $\mathcal{B}$ by employing negatives $I_n^* = \argmin_{I_n \in \mathcal{B}: d_{an} > d_{ap}} d_{an}$. Based on this idea, different online\cite{Parkhi15,npairs} and offline\cite{smartmining} strategies emerged.
\\
\textbf{(Static) Distance-based sampling:}
By considering the hardness of a negative, one can successfully discard easy and uninformative triplets. However, triplets that are too hard lead to noisy learning signals due to overall high gradient variance\cite{margin}. As a remedy, to control the variance while maintaining sufficient triplet utility, sampling can be extended to also consider easier negatives, i.e. introducing a sampling distribution $I_n \sim p(I_n|I_a)$ over the range of distances $d_{an}$ between anchor and negatives. Wu \etal\cite{margin} propose to sample from a static uniform prior on the range of $d_{an}$, thus equally considering negatives from the whole spectrum of difficulties. As pairwise distances on $\Phi$ are strongly biased towards larger $d_{an}$, their sampling distribution requires to weigh $p(I_n|I_a)$ inversely to the analytical distance distribution on $\Phi$: $q(d) \propto d^{D-2}\left[ 1 - \frac{1}{4}d^2 \right]^{\frac{D-3}{2}}$ for large $D \geq 128$\cite{p_hypersphere}. Distance-based sampling from the static, uniform prior is then performed by

\begin{equation}
I_n \sim p(I_n|I_a) \propto \min\left(\lambda, q^{-1}(d_{an})\right)
\label{eq:hard_neg}
\end{equation}

\noindent
with $\lambda$ being a clipping hyperparameter for regularization.


\section{Learning an Adaptive Negative Sampling}

\begin{figure*}[t]
\begin{center}
\includegraphics[width=0.99\textwidth]{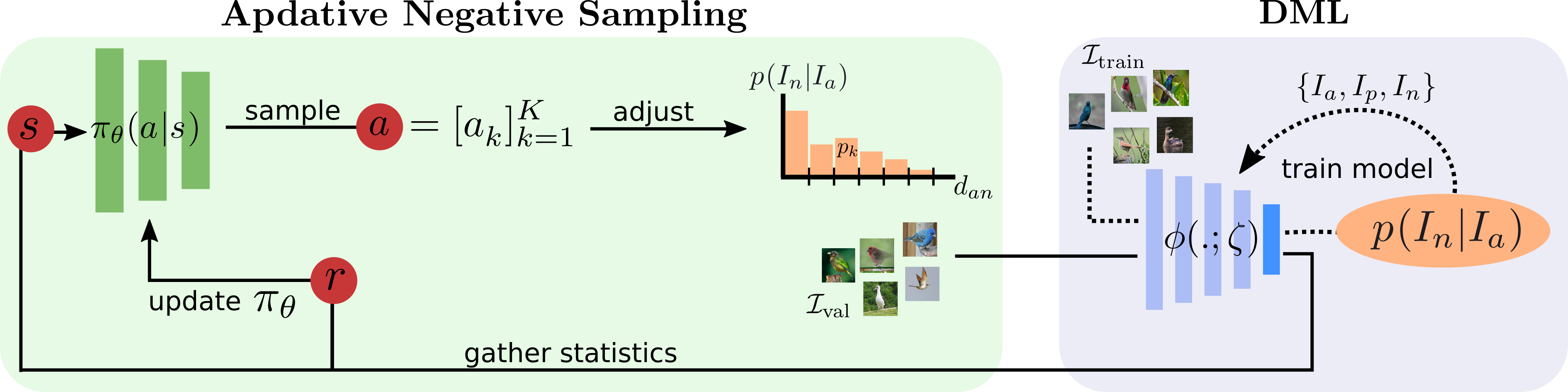}
\end{center}
  \caption{\textit{Overview of approach.} Blue denotes the standard Deep Metric Learning (DML) setup using triplets $\{I_a, I_p, I_n\}$. Our proposed adaptive negative sampling is shown in green: (1) We compute the current training state $s$ using $\mathcal{I}_{val}$. (2) Conditioned on $s$, our policy $\pi_{\theta}(a|s)$ predicts adjustments to $p_k$. (3) We perform bin-wise adjustments of $p(I_n|I_a)$. (4) Using the adjusted $p(I_n|I_a)$ we train the DML model. (5) Finally, $\pi_\theta$ is updated based on the reward $r$.}
\label{fig:learning_overview}
\end{figure*}

Distance-based sampling of negatives $I_n$ has proven to offer a good trade-off between fast convergence and a stable, informative training signal. However, a static sampling distribution $p(I_n|I_a)$ provides a stream of training data independent of the the changing needs of a DML model during learning. While samples of mixed difficulty may be useful at the beginning, later training stages are calling for samples of increased difficulty, as e.g. analyzed by curriculum learning\cite{curriculum_learning}. Unfortunately, as different models and even different model intializations\cite{GlorotB10} exhibit distinct learning dynamics, finding a generally applicable learning schedule is challenging. Thus, again, heuristics\cite{CL_icml} are typically employed, inferring changes after a fixed number of training epochs or iterations. To provide an optimal training signal, however, we rather want $p(I_n|I_a)$ to adapt to the training state of the DML model than merely the training iteration. Such an adaptive negative sampling allows for adjustments which directly facilitate maximal DML performance. Since manually designing such a strategy is difficult, \textit{learning} it is the most viable option. 
\\
Subsequently, we first present how to find a parametrization of $p(I_a|I_n)$ that is able to represent arbitrary, potentially multi-modal distributions, thus being able to sample negatives $I_n$ of any mixture of difficulty needed. Using this, we can learn a policy which effectively alters $p(I_n|I_a)$ to optimally support learning of the DML model.

\subsection{Modelling a flexible sampling distribution}
\label{sec:model_p}
Since learning benefits from a diverse distribution $p(I_n|I_a)$ of negatives, uni-modal distributions 
(e.g. Gaussians, Binomials, $\chi^2$) are insufficient. Thus, we utilize a discrete probability mass function $p(I_n|I_a) := Pr\{d_{an} \in u_k\} = p_k$, where the bounded intervall $U = [\lambda_{\text{min}}, \lambda_{\text{max}}]$ of possible distances $d_{an}$ is discretized into disjoint equidistant bins $u_1,\dots,u_K$. The probability of drawing $I_n$ from bin $u_k$ is $p_k$ with $p_k \geq 0$ and $\sum_k p_k = 1$. Fig. \ref{fig:model_sampling_distr} illustrates this discretized sampling distribution. 
\\
This representation of the negative sampling distribution effectively controls \textit{which} samples are used to learn $\phi$. As $\phi$ changes during learning, $p(I_n|I_a)$ should also adapt to always provide the most useful training samples, i.e. to control \textit{when} to use \textit{which} sample. Hence the probabilities $p_k$ need to be updated while learning $\phi$. We subsequently solve this task by learning a stochastic adjustment policy $\pi_\theta$ for the $p_k$, implemented as a neural network parametrized by $\theta$.

\subsection{Learning an adjustment policy for $p(I_n|I_a)$}
Our sampling process based on $p(I_n|I_a)$ should provide optimal training signals for learning $\phi$ at every stage of training. Thus, we adjust the $p_k$ by a multiplicative update $a \in \mathcal{A}$ conditioned on the current representation (or state) $s \in \mathcal{S}$ of $\phi$ during learning. We introduce a conditional distribution $\pi_\theta(a|s)$ to control which adjustment to apply at which state $s$ of training $\phi$. To learn $\pi_\theta$, we measure the utility of these adjustments for learning $\phi$ using a reward signal $r=r(s,a)$. We now first describe how to model each of these components, before presenting how to efficiently optimize the adjustment policy $\pi_\theta$ alongside $\phi$. 
\\
\textbf{Adjustments $a$:} To adjust $p(I_n|I_a)$, $\pi_\theta(a|s)$ proposes adjustments $a$ to the $p_k$. To lower the complexity of the action space, we use a limited set of actions $\mathcal{A} = \{\alpha, 1, \beta\}$ to individually decrease, maintain, or increase the probabilities $p_k$ for each bin $u_k$, i.e. $a:=[a_k \in \{\alpha, 1, \beta\}]_{k=1}^K$. Further, $\alpha, \beta$ are fixed constants $0<\alpha<1, \beta> 1$ and $\frac{\alpha+\beta}{2}=1$. Updating $p(I_n|I_a)$ is then simply performed by bin-wise updates $p_k \leftarrow p_k \cdot a_k$ followed by re-normalization. Using a multiplicative adjustment accounts for the exponential distribution of distances on $\Phi$ (cf. Sec. \ref{sec:static_dist_sampling}).
\\
\textbf{Training states $s$:} Adjustments $a$ depend on the present state $s \in \mathcal{S}$ of the representation $\phi$. Unfortunately, we cannot use the current model weights $\zeta$ of the embedding network, as the dimensionality of $s$ would be to high, thus making optimization of $\pi_\theta$ infeasible. Instead, we represent the current training state using representative statistics describing the learning progress: running averages over Recall@1\cite{recall}, NMI\cite{nmi} and average distances between and within classes on a fixed held-back validation set $\mathcal{I}_{\text{val}}$ . Additionally we use past parametrizations of $p(I_n|I_a)$ and the relative training iteration (cf. Implementation details, Sec. \ref{sec:exps}).
\\
\textbf{Rewards $r$:} 
An optimal sampling distribution $p(I_n|I_a)$ yields triplets whose training signal consistently improves the evaluation performance of $\phi$ while learning. Thus, we compute the reward $r$ for 
for adjustments $a \sim \pi_\theta(a|s)$ 
by directly measuring the relative improvement of $\phi(\cdot;\zeta)$ over $\phi(\cdot;\zeta^\prime)$ from the previous training state. This improvement is quantified through DML evaluation metrics $e(\phi(.;\zeta_t), \mathcal{I}_{\text{val}})$ on the validation set $\mathcal{I}_{\text{val}}$. More precisely, we define $r$ as


\begin{equation}
r = \text{sign}\left(e(\phi(.;\zeta), \mathcal{I}_{\text{val}})- e(\phi(.;\zeta^\prime), \mathcal{I}_{\text{val}}) )\right)
\label{eq:reward}
\end{equation}

\noindent
where $\zeta$ was reached from $\zeta^\prime$ after $M$ DML training iterations using $p(I_n|I_a)$. We choose $e$ to be the sum of Recall@1\cite{recall} and NMI\cite{nmi}. Both metrics are in the range $[0,1]$ and target slightly different performance aspects. Further, similar to \cite{ala}, we utilize the sign function for consistent learning signals even during saturated training stages.
\\
\textbf{Learning of $\pi_\theta$:} 
Adjusting $p(I_n|I_a)$ is a stochastic process controlled by actions $a$ sampled from $\pi_\theta(a|s)$ based on a current state $s$. This defines a Markov Decision Process (MDP) naturally optimized by Reinforcement Learning. The policy objective $J(\theta)$ is formulated to maximize the total expected reward $R(\tau) = \sum_{t} r_t(a_t,s_t)$ over training episodes of tuples $\tau = \{(a_t, s_t, r_t) | t = 0,\dots,T]\}$ collected from sequences of $T$ time-steps, i.e.

\begin{equation}
J(\theta) = \mathbb{E}_{\tau \sim \pi_{\theta}(\tau)}[R(\tau)]
\label{eq:policy_objective}
\end{equation}

\noindent
Hence, $\pi_\theta$ is optimized to predict adjustments $a$ for $p(I_n|I_a)$ which yield high rewards and thereby improving the performance of $\phi$. 
Common approaches use episodes $\tau$ comprising long state trajectories which potentially cover multiple training epochs\cite{neural_data_filter}. As a result, there is a large temporal discrepancy between model and policy updates.
However, in order to closely adapt $p(I_n|I_a)$ to the learning of $\phi$, this discrepancy needs to be minimized. 
In fact, our experiments show that single-step episodes, i.e. $T=1$, are sufficient for optimizing $\pi_\theta$ to infer meaningful adjustments $a$ for $p(I_n|I_a)$. Such a setup is also successfully adopted by contextual bandits \cite{cbandits} \footnote{Opposed to bandits, in our RL setup, actions which are sampled from $\pi_\theta$ influence \textit{future} training states of the learner. Thus, the policy implicitly learns state-transition dynamics.}. In summary, our training episodes $\tau$ consists of updating $p(I_n|I_a)$ using a sampled adjustment $a$, performing $M$ DML training iterations based on the adjusted $p(I_n|I_a)$ and updating $\pi_\theta$ using the resulting reward $r$. Optimizing Eq. \ref{eq:policy_objective} is then performed by standard RL algorithms which approximate different variations of the policy gradient based on the gain $G(s,a)$,


\begin{equation}
\nabla_{\theta} J(\theta) = \mathbb{E}_{\tau \sim \pi_{\theta}(\tau)}\left[\nabla_{\theta} \log \pi_{\theta}(a|s) G(s,a)\right]
\label{eq:policy_gradient}
\end{equation}

\noindent
The choice of the exact form of $G = G(s,a)$ gives rise to different optimization methods, e.g REINFORCE\cite{reinforce} ($G = R(\tau)$), Advantage Actor Critic (A2C)\cite{Sutton1998} ($G = A(s,a)$), etc. Other RL algorithms, such as TRPO\cite{trpo} or PPO\cite{ppo} replace Eq. \ref{eq:policy_objective} by surrogate objective functions. Fig. \ref{fig:learning_overview} provides an overview over the learning procedure. Moreover, in the supplementary material we compare different RL algorithms and summarizes the learning procedure in Alg. 1 using PPO\cite{ppo} for policy optimization.
\\
\textbf{Initialization of $p(I_n|I_a)$:} We find that an initialization with a slight emphasis towards smaller distances $d_{an}$ works best. However, as shown in Tab. \ref{tab:ablation}, also other initializations work well. In addition, the limits of the distance interval $U=[\lambda_{\text{min}}, \lambda_{\text{max}}]$ can be controlled for additional regularization as done in \cite{margin}. This means ignoring values above $\lambda_\text{max}$ and clipping values below $\lambda_\text{min}$, which is analysed in Tab. \ref{tab:ablation}.
\\
\textbf{Self-Regularisation}: As noted in \cite{mic}, the utilisation of intra-class features can be beneficial to generalization. Our approach easily allows for a learnable inclusion of such features. As positive samples are generally closest to anchors, we can merge positive samples into the set of negative samples and have the policy learn to place higher sampling probability on such low-distance cases. We find that this additionally improves generalization performance.
\\
\textbf{Computational costs:}
Computational overhead over fixed sampling strategies\cite{semihard,margin} comes from the estimation of $r$ requiring a forward pass over $\mathcal{I}_{\text{val}}$ and the computation of the evaluation metrics. For example, setting $M=30$ increases the computation time per epoch by less than $20\%$.




 
 \begin{table*}[t]
   \label{tab:baselines}
   \setlength\tabcolsep{1.4pt}
   \centering
   \begin{tabular}{l|c|cccc|cccc|cccc}
     \toprule
     \multicolumn{2}{l}{Dataset} & \multicolumn{4}{c}{CUB200-2011\cite{cub200-2011}} & \multicolumn{4}{c}{CARS196\cite{cars196}} & \multicolumn{4}{c}{SOP\cite{lifted}} \\
     \midrule
     Approach & Dim & R@1 & R@2 & R@4 & NMI & R@1 & R@2 & R@4 & NMI & R@1 & R@10 & R@100 & NMI\\
     \midrule
     Margin\cite{margin} + $\mathcal{U}$-dist (orig) & 128 & 63.6 & 74.4 & 83.1 & 69.0 & 79.6 & 86.5 & 90.1 & \textbf{69.1} & 72.7 & 86.2 & 93.8 & \textbf{90.7} \\
     Margin\cite{margin} + $\mathcal{U}$-dist (ReImp, $\beta=1.2$) & 128 & 63.5 & 74.9 & 84.4 & 68.1 & 80.1 & 87.4 & 91.9 & 67.6 & 74.6 & 87.5 & 94.2 & 90.7 \\
    Margin\cite{margin} + $\mathcal{U}$-dist (ReImp, $\beta=0.6$) & 128 & 63.0 & 74.3 & 83.0 & 66.9 & 79.7 & 87.0 & 91.8 & 67.1 & 73.5 & 87.2 & 93.9 & 89.3 \\
     \textbf{Margin\cite{margin} + \textit{PADS} (Ours)} & 128 & \textbf{67.3} & \textbf{78.0} & \textbf{85.9} & \textbf{69.9} & \textbf{83.5} & \textbf{89.7} & \textbf{93.8} & 68.8 & \textbf{76.5} & \textbf{89.0} & \textbf{95.4} & 89.9\\
     \hline
      \hline
     Triplet\cite{semihard} + semihard (orig) & 64 & 42.6 & 55.0 & 66.4 & 55.4 & 51.5 & 63.8 & 73.5 & 53.4 & 66.7 & 82.4 & 91.9 & \textbf{89.5} \\
     Triplet\cite{semihard} + semihard (ReImp) & 128 & 60.6 & 72.3 & 82.1 & 65.5 & 71.9 & 81.5 & 88.5 & 64.1 & 73.5 & 87.5 & 94.9 & 89.2 \\
     Triplet\cite{semihard} + $\mathcal{U}$-dist (ReImp) & 128 & 62.2 & 73.2 & 82.8 & 66.3 & 78.0 & 85.6 & 91.4 & 65.7 & 73.9 & 87.7 & 94.5 & 89.3\\
     \textbf{Triplet\cite{semihard} + \textit{PADS} (Ours)} & 128 & \textbf{64.0} & \textbf{75.5} & \textbf{84.3} & \textbf{67.8} & \textbf{79.9} & \textbf{87.5} & \textbf{92.3} & \textbf{67.1} & \textbf{74.8} & \textbf{88.2} & \textbf{95.0} & \textbf{89.5}\\
     \bottomrule
   \end{tabular}
     \caption{Comparison of our proposed adaptive negative sampling (\textit{PADS}) against common static negative sampling strategies: semihard negative mining\cite{lifted} (\emph{semihard}) and static distance-based sampling (\emph{$\mathcal{U}$-dist})\cite{margin} using triplet\cite{semihard} and margin loss\cite{margin}. \emph{ReImp.} denotes our re-implementations and \emph{Dim} the dimensionality of $\phi$.} 
     \label{tab:baselines}
 \end{table*}

\section{Experiments}
\label{sec:exps}
In this section we provide implementation details, evaluations on standard metric learning datasets, ablations studies and analysis experiments.
\\
\textbf{Implementation details.} We follow the training protocol of \cite{margin} with ResNet50. During training, images are resized to $256\times 256$ with random crop to $224\times 224$ and random horizontal flipping. For completeness, we also evaluate on Inception-BN \cite{googlenetv2} following standard practice in the supplementary. The initial learning rates are set to $10^{-5}$. We choose triplet parameters according to \cite{margin}, with $\gamma = 0.2$. For margin loss, we evaluate margins $\beta=0.6$ and $\beta=1.2$. Our policy $\pi$ is implemented as a two-layer fully-connected network with ReLU-nonlinearity inbetween and 128 neurons per layer. Action values are set to $\alpha=0.8, \beta=1.25$. Episode iterations $M$ are determined via cross-validation within [30,150]. The sampling range $[\lambda_{\text{min}}, \lambda_{\text{min}}]$ of $p(I_n|I_a)$ is set to [0.1, 1.4], with $K=30$. The sampling probability of negatives corresponding to distances outside this interval is set to $0$. For the input state we use running averages of validation recall, NMI and average intra- and interclass distance based on running average lengths of 2, 8, 16 and 32 to account for short- and longterm changes. We also incorporate the metrics of the previous 20 iterations. Finally, we include the sampling distributions of the previous iteration and the training progress normalized over the total training length. For optimization, we utilize an A2C + PPO setup with ratio limit $\epsilon=0.2$. The history policy is updated every 5 policy iterations. For implementation we use the PyTorch framework\cite{pytorch} on a single NVIDIA Titan X. 
\\
\textbf{Benchmark datasets.}
We evaluate the performance on three common benchmark datasets.
For each dataset the first half of classes is used for training and the other half is used for testing. Further, we use a random subset of $15\%$ of the training images for our validation set $\mathcal{I}_{\text{val}}$. We use: \\
\textit{CARS196}\cite{cars196}, with 16,185 images from 196 car classes. \\
\textit{CUB200-2011}\cite{cub200-2011}, 11,788 bird images from 200 classes.\\
\textit{Stanford Online Products (SOP)}\cite{lifted}, containing 120,053 images divided in 22,634 classes.\\

\begin{figure}[b]
\begin{center}
\includegraphics[width=0.44\textwidth]{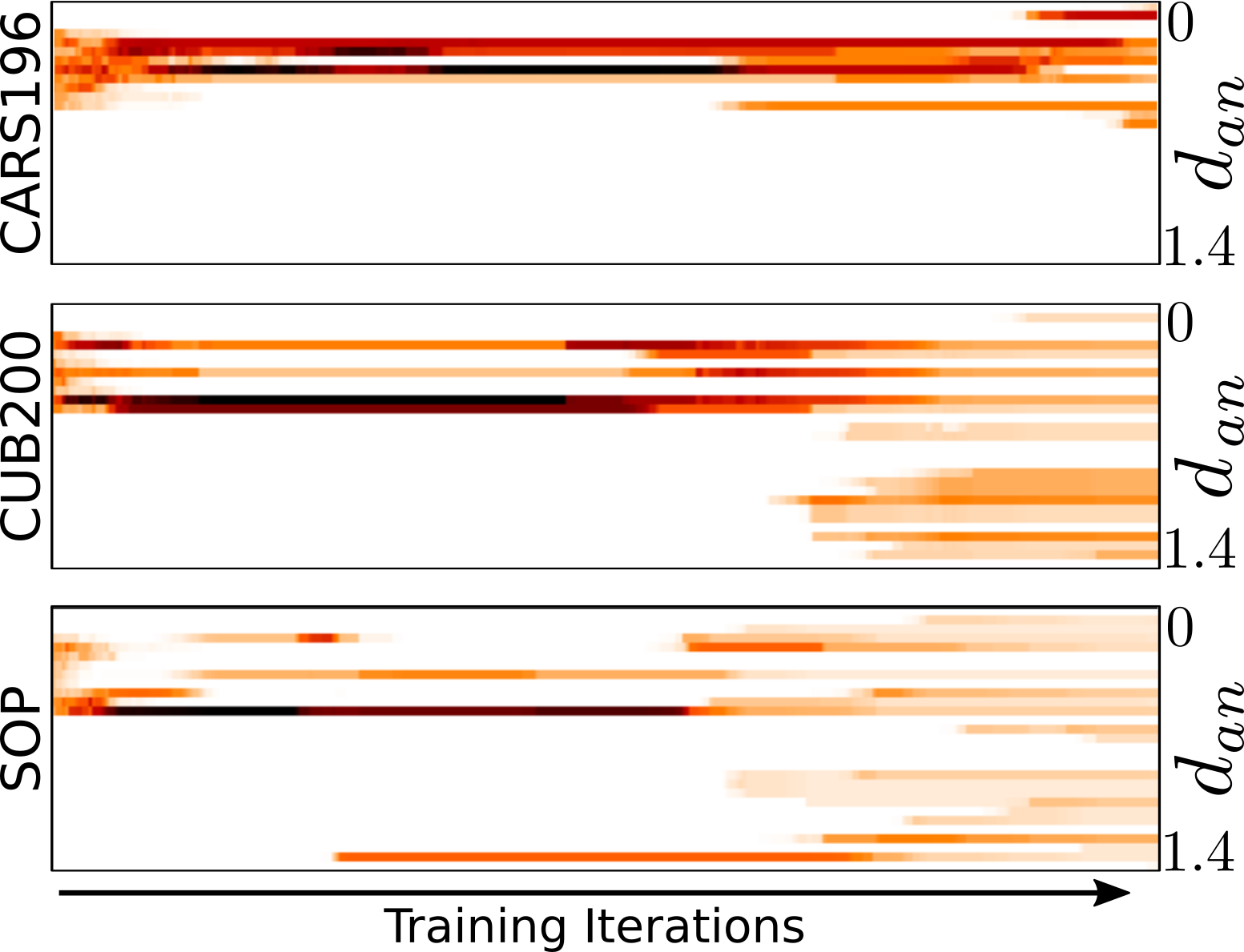}
\end{center}
   \caption{\textit{Averaged progression of $p(I_n|I_a)$ over multiple training runs on CUB200-2011, CARS196 and SOP}.}
\label{fig:p_dist_progression}
\end{figure}

\subsection{Results}
In Tab. \ref{tab:baselines} we apply our adaptive sampling strategy on two widely adopted basic ranking losses: triplet\cite{semihard} and margin loss\cite{margin}. For each loss, we compare against the most commonly used static sampling strategies, semi-hard\cite{semihard} (semihard) and distance-based sampling\cite{margin} ($\mathcal{U}$-dist) on the CUB200-2011, CARS196 and SOP dataset. We measure image retrieval performance using recall accuracy R@k\cite{recall} following \cite{abier}. For completeness we additonally show the normalized mutual information score (NMI)\cite{nmi}, despite not fully correlating with retrieval performance. For both losses and each dataset, our learned negative sampling significantly improves the performance over the non-adaptive sampling strategies. Especially the strong margin loss greatly benefits from the adaptive sampling, resulting in boosts up to $3.8\%$ on CUB200-2011, $3.4\%$ on CARS196 and $1.9\%$ on SOP. This clearly demonstrates the importance of adjusting triplet sampling to the learning process a DML model, especially for smaller datasets.
\\
Next, we compare these results with the current state-of-the-art in DML which extend these basic losses using diverse additional training signals (MIC\cite{mic}, DVML\cite{dvml}, HORDE\cite{horde}, A-BIER\cite{abier}), ensembles of embedding spaces (DREML\cite{dreml}, D\&C\cite{Sanakoyeu_2019_CVPR}, Rank\cite{rankedlist}) and/or significantly more network parameters (HORDE\cite{horde}, SOFT-TRIPLE\cite{softriple}). Tab. \ref{tab:sota} shows that our results, despite not using such additional extensions, compete and partly even surpass these strong methods. On CUB200-2011 we outperform all methods, including the powerful ensembles, by at least $1.2\%$ in Recall accuracy. On CARS196\cite{cars196} we rank second behind the top performing non-ensemble method D\&C\cite{Sanakoyeu_2019_CVPR}. On SOP\cite{lifted} we lose $0.7\%$ to MIC\cite{mic} which, in turn, we surpass on both CUB200-2011 and CARS196. This highlights the strong benefit of our adaptive sampling.

\begin{table*}[t]
    \setlength\tabcolsep{1.5pt}
    \centering
     \begin{tabular}{l|c|ccc|c||ccc|c||ccc|c}
        \toprule
        Dataset & & \multicolumn{4}{c}{CUB200-2011\cite{cub200-2011}} & \multicolumn{4}{c}{CARS196\cite{cars196}}
        & \multicolumn{4}{c}{SOP\cite{lifted}}\\ 
        \hline
        Approach & Dim & R@1 & R@2 & R@4  & NMI & R@1 & R@2 & R@4  & NMI & R@1 & R@2 & R@4  & NMI \\
        \midrule
        HTG\cite{htg} & 512 & 59.5 & 71.8 & 81.3 & -                                  & 76.5 & 84.7 & 90.4 & - 
                        & - & - & - & -\\
        HDML\cite{hardness-aware} & 512 & 53.7 & 65.7 & 76.7 & 62.6                 & 79.1 & 87.1 & 92.1 & 69.7
                        & 68.7 & 83.2 & 92.4 & 89.3\\
        HTL\cite{htl} & 512 & 57.1 & 68.8 & 78.7 & -                                & 81.4 & 88.0 & 92.7  & - 
                        & 74.8 & 88.3 & 94.8  & -\\
        DVML\cite{dvml} & 512 & 52.7 & 65.1 & 75.5 & 61.4                           & 82.0 & 88.4 & 93.3 & 67.6
                        & 70.2 & 85.2 & 93.8 & 90.8\\
        A-BIER\cite{abier} & 512 & 57.5 & 68.7 & 78.3 & -                           & 82.0 & 89.0 & 93.2  & - 
                        & 74.2 & 86.9 & 94.0  & -\\
        MIC\cite{mic} & 128 & 66.1 & 76.8 & 85.6 & 69.7                             & 82.6 & 89.1 & 93.2 & 68.4 
                        & 77.2 & 89.4 & 95.6 & 90.0\\
        D\&C\cite{Sanakoyeu_2019_CVPR} & 128 & 65.9 & 76.6 & 84.4 & 69.6            & 84.6 & 90.7 & 94.1 & 70.3
                            & 75.9 & 88.4 & 94.9 & 90.2\\
        Margin\cite{margin} & 128 & 63.6 & 74.4 & 83.1 & 69.0                     & 79.6 & 86.5 & 90.1 & 69.1
                        &  72.7 & 86.2 & 93.8 & 90.8\\                            
        \textbf{Ours (Margin\cite{margin} + \textit{PADS})} & 128 & \textbf{67.3} & \textbf{78.0} & \textbf{85.9} & \textbf{69.9} & 83.5 & 89.7 & 93.8 & 68.8 
                        & 76.5 & 89.0 & 95.4 & 89.9\\                        
        \hline
        \multicolumn{14}{l}{Significant increase in network parameter:} \\      
        \hline
        HORDE\cite{horde}+contrastive loss\cite{contrastive} & 512 & 66.3 & 76.7 & 84.7 & -                   & 83.9 & 90.3 & 94.1 & - 
                        & - & - & - & -\\
        SOFT-TRIPLE\cite{softriple} & 512 & 65.4 & 76.4 & 84.5& -             & 84.5 & 90.7 & 94.5 & 70.1
                        & 78.3 & 90.3 & 95.9 & \textbf{92.0}\\
        \hline
        \multicolumn{14}{l}{Ensemble Methods:} \\
        \hline
        Rank\cite{rankedlist} & 1536 & 61.3 & 72.7 & 82.7 & 66.1               & 82.1 & 89.3 & 93.7& 71.8
                        & \textbf{79.8} & \textbf{91.3} & \textbf{96.3} & 90.4\\
        DREML\cite{dreml} & 9216 & 63.9 & 75.0 & 83.1 & 67.8                        & \textbf{86.0} & \textbf{91.7} & \textbf{95.0} & \textbf{76.4} 
                        & - & - & - & -\\
        ABE\cite{abe} & 512 & 60.6 & 71.5 & 79.8 & -                                & 85.2 & 90.5 & 94.0 & - 
                        & 76.3 & 88.4 & 94.8 & -\\
        \bottomrule
    \end{tabular}
    \caption{Comparison to the state-of-the-art DML methods on CUB200-2011\cite{cub200-2011}, CARS196\cite{cars196} and SOP\cite{lifted}. \emph{Dim} denotes the dimensionality of $\phi$.}
    \label{tab:sota}
\end{table*}

\subsection{Analysis}
We now present various analysis experiments providing detailed insights into our learned adaptive sampling strategy. 
\\
\noindent
\textbf{Training progression of $p(I_n|I_a)$:}
We now analyze in Fig. \ref{fig:p_dist_progression} how our adaptive sampling distribution progresses during training by averaging the results of multiple training runs with different network initializations. While on CARS196 the distribution $p(I_n|I_a)$ strongly emphasizes smaller distances $d_{an}$, we observe on CUB200-2011 and SOP generally a larger variance of $p(I_n|I_a)$. Further, on each dataset, during the first half of training $p(I_n|I_a)$ quickly peaks on a sparse set of bins $u_k$, as intuitively expected, since most triplets are still informative. As training continues, $p(I_n|I_a)$ begins to yield both harder and easier negatives, thus effectively sampling from a wider distribution. This observation confirms the result of Wu \etal\cite{margin} which proposes to ease the large gradient variance introduced by hard negatives with also adding easier negatives. Moreover, for each dataset we observe a different progression of $p(I_n|I_a)$ which indicates that manually designing similar sampling strategies is difficult, as also confirmed by our results in Tab. \ref{tab:baselines} and \ref{tab:curriculum_learning}.\\

\begin{table}[b]
\begin{subtable}{0.5\textwidth}
  \centering
  \begin{tabular}{l|c|c|c|c}
    \toprule
     & Init. & Reference & fix $\pi_\theta$  & fix last $p(I_n|I_a)$ \\
    \midrule
    R@1 & $\neq$ & \textbf{65.4} & 64.3 & 59.0 \\
    \hline
    R@1 & = & 65.4 & \textbf{65.8} & 57.6 \\    
    \bottomrule
\end{tabular}
\end{subtable}
\caption{Transferring a fixed trained policy $\pi_\theta$ and fixed final distribution $p(I_n|I_a)$ to training runs with different ($\neq$) and the same network initialization (=). \emph{Reference} denotes the training run from which $\pi_\theta$ and $p(I_n|I_a)$ is obtained.}
\label{tab:policy_transfer}
\vspace{-3mm}
\end{table}

\begin{table}[b]
\begin{subtable}{0.5\textwidth}
  \centering
  \begin{tabular}{l|cc|cc}
    \toprule
     Dataset & \multicolumn{2}{c}{CUB200-2011\cite{cub200-2011}} & \multicolumn{2}{c}{CARS196\cite{cars196}}  \\
     \midrule
      Metrics & R@1 & NMI & R@1 & NMI \\
    \midrule
   \textbf{Ours} & \textbf{67.3} & \textbf{69.9} & \textbf{83.5} & \textbf{68.8} \\
    linear CL & 59.1 & 63.1 & 72.2 & 64.0\\    
    non-linear CL & 63.6 & 68.4 & 78.1 & 66.8 \\ 
    \bottomrule
\end{tabular}
\end{subtable}
\caption{Comparison to curriculum learning strategies with predefined linear and non-linear progression of $p(I_n|I_a)$.}
\label{tab:curriculum_learning}
\vspace{-3mm}
\end{table}

\noindent
\textbf{Transfer of $\pi_\theta$ and $p(I_n|I_a)$:}
Tab. \ref{tab:policy_transfer} investigates how well a trained policy $\pi_\theta$ or final sampling distribution $p(I_n|I_a)$ from a reference run transfer to differently (\emph{$\neq$}) or equally (\emph{$=$}) initialized training runs.
We find that applying a fixed trained policy (\emph{fix $\pi_\theta$}) to a new training run with the same network initialization (\emph{$=$}) improves performance by $0.4\%$ due to the immediate utility of $\pi_\theta$ for learning $\phi$ as $\pi_\theta$ is already fully adapted to the reference learning process. In contrast, applying the trained policy to a differently initialized training run ($\neq$) drops performance by $1.5\%$. Since the fixed $\pi_\theta$ cannot adapt to the learning states of the new model, its support for optimizing $\phi$ is diminished. Note that the policy has only been trained on a single training run, thus it cannot fully generalize to different training dynamics. This shows the importance of an adaptive sampling.
\\
Next, we investigate if the distribution $p(I_n|I_a)$ obtained at the end of training can be regarded as an optimal sampling distribution over $d_{an}$, as $\pi_\theta$ is fully trained. To this end we fix and apply the distribution $p(I_n|I_a)$ after its last adjustment by $\pi_\theta$ (\emph{fix last $p(I_n|I_a)$}) in training the reference run. As intuitively expected, in both cases performance drops strongly as \emph{(i)} we now have a static sampling process and \emph{(ii)} the sampling distribution is optimized to a specific training state. Given our strong results, this proves that our sampling process indeed adapts to the learning of $\phi$.\\

\noindent
\textbf{Curriculum Learning:}
To compare our adaptive sampling with basic curriculum learning strategies, we pre-define two sampling schedules: \textit{(1)} A \emph{linear} increase of negative hardness, starting from a semi-hard distance intervall\cite{semihard} and \textit{(2)} a \textit{non-linear} schedule using distance-based sampling\cite{margin}, where the distribution is gradually shifted towards harder negatives. We visualize the corresponding progression of the sampling distribution in the supplementary material. 
Tab. \ref{tab:curriculum_learning} illustrates that both fixed, pre-defined curriculum schedules perform worse than our learned, adaptive sampling distribution by at least $3.6\%$ on CUB200-2011. On CARS196 the performance gap is even larger. The strong difference in datasets further demonstrates the difficulty of finding broadly applicable, effective fixed sampling strategies.  
%

\begin{table}
\centering

\begin{subtable}{0.5\textwidth}
  \centering
  \begin{tabular}{l|c|c|c|c}
    \toprule
    \textbf{$[\lambda_{\text{min}},\lambda_{\text{max}}]$} & $[0,2]$ & $[0.1,1.4]$ & $[0.25,1.0]$ & $[0.5,1.4]$\\
    \midrule
    Recall@1 & $64.7$ & $\mathbf{65.7}$ & $64.8$ & $63.7$\\  
    NMI      & $67.5$ & $\mathbf{69.2}$ & $68.2$ & $67.5$\\   
    \bottomrule
  \end{tabular}
\caption{Varying the interval $U=[\lambda_{\text{min}}, \lambda_{\max}]$ of distances $d_{an}$ used for learning $p(I_n|I_a)$. The number of bins $u_k$ is kept fixed to $K=30$.}
\end{subtable}

\bigskip
\begin{subtable}{0.5\textwidth}
  \centering
  \begin{tabular}{l|c|c|c|c}
    \toprule
    \textbf{Num. bins $K$} & $10$ & $30$ & $50$ & $100$\\
    \midrule
    Recall@1  & $63.8$ & $\mathbf{65.7}$ & $65.3$ & $64.9$\\
    NMI       & $67.8$ & $\mathbf{69.2}$ & $68.7$ & $68.6$\\     
    \bottomrule
  \end{tabular}
\caption{Varying the number of bins $u_k$ used to discretize the range of distances $U = [0.1,1.4]$ used for learning $p(I_n|I_a)$.}
\end{subtable}

\bigskip
\begin{subtable}{0.5\textwidth}
  \centering
  \begin{tabular}{l|c|c|c}
    \toprule
    \textbf{Init. Distr.} & $\mathcal{U}_{[0.1,1.4]}$ & $\mathcal{N}(0.5, 0.05)$ & $\mathcal{U}_{[0.3,0.7]}$ \\
    \midrule
    Recall@1 & $63.9$ & $65.0$ & $\mathbf{65.7}$ \\
    NMI      & $67.0$ & $68.6$ & $\mathbf{69.2}$ \\
    \bottomrule
  \end{tabular}
\caption{Comparison of $p(I_n|I_a)$-initializations on distance interval $U=[0.1,1.4]$. $\mathcal{U}_{[a,b]}$ denotes uniform emphasis in $[a,b]$ with low probabilities outside the interval. $\mathcal{N}(\mu, \sigma)$ denotes a normal distribution.}
\end{subtable}

\caption{Ablation experiments analyzing various parameters for learning $p(I_n|I_a)$.}
\label{tab:ablation}
\vspace{-2mm}
\end{table}

\subsection{Ablation studies}
\label{sec:Ablations}
Subsequently we ablate different parameters for learning our sampling distribution $p(I_n|I_a)$ on the CUB200-2011 dataset. More ablations are shown in the appendix. To make the following experiments comparable, no learning rate scheduling was applied, as convergence may significantly change with different parameter settings. In contrast, the results in Tab \ref{tab:baselines}-\ref{tab:sota} are obtained with our best parameter settings \textit{and} a fixed learning rate scheduling. Without scheduling, our best parameter setting achieves a recall value of $65.7$ and NMI of $69.2$ on CUB200-2011.
\\

\noindent
\textbf{Distance interval $U$:} As presented in Sec. \ref{sec:model_p}, $\tilde{p}(I_n|I_a)$ is defined on a fixed interval $U=[\lambda_{\text{min}}, \lambda_{\text{max}}]$ of distances. Similar to other works\cite{margin,smartmining}, this allows us to additionally regularize the sampling process by clipping the tails of the true range of distances $[0,2]$ on $\Phi$. Tab. \ref{tab:ablation} (a) compares different combinations of $\lambda_{\text{min}}, \lambda_{\text{max}}$. We observe that, while each option leads to significant performance boost compared to the static sampling strategies, an interval $U=[0.1,1.4]$ results in the most effective sampling process. 
\\

\noindent
\textbf{Number of bins $K$:} Next, we analyze the impact of the $U$ resolution in Tab. \ref{tab:ablation} (b), i.e. the number of bins $K$. This affects the flexibility of $p(I_n|I_a)$, but also the complexity of the actions $a$ to be predicted. As intuitively expected, increasing $K$ allows for better adaption and performance until the complexity grows too large.
\\

\noindent
\textbf{Initialization of $p(I_n|I_a)$:} Finally, we analyze how the initialization of $p(I_n|I_a)$ impacts learning. Tab. \ref{tab:ablation} (c) compares the performance using different initial distributions, such as a neutral uniform initialization (i.e. random sampling) ($\mathcal{U}_{[0.1,1.4]}$), emphasizing semi-hard negatives $I_n$ early on ($\mathcal{U}_{[0.3,0.7]}$) or a proxy to \cite{margin} ($\mathcal{N}(0.5,0.05)$). We observe that our learned sampling process benefits from a meaningful, but generic initial configuration of $p(I_n|I_a)$, $\mathcal{U}_{[0.3,0.7]}$, to effectively adapt the learning process of $\phi$.


\section{Conclusion}
This paper presents a learned adaptive triplet sampling strategy using Reinforcement Learning. We optimize a teacher network to adjust the negative sampling distribution to the ongoing training state of a DML model. By training the teacher to directly improve the evaluation metric on a held-back validation set, the resulting training signal optimally facilitates DML learning. Our experiments show that our adaptive sampling strategy improves significantly over static sampling distributions. Thus, even though only built on top of basic triplet losses, we achieve competitive or even superior performance compared to the state-of-the-art of DML on multiple standard benchmarks sets.

\section*{Acknowledgements}
We thank David Yu-Tung Hui (MILA) for valuable insights regarding the choice of RL Methods. This work has been supported in part by Bayer AG, the German federal ministry BMWi within the project “KI Absicherung”, and a hardware donation from NVIDIA corporation.

{\small
\bibliographystyle{ieee_fullname}
\bibliography{egbib}
}

\clearpage
\newpage
\appendix
\section*{Supplementary Material}
This part contains supporting or additional experiments to the main paper, such as additional ablations and qualitative evaluations.

\section{Additional Ablation Experiments}
We now conduct further ablation experiments for different aspects of our proposed approach based on the CUB200-2011\cite{cub200-2011} dataset. Note, that like in our main paper we did not apply any learning rate scheduling for the results of our approach to establish comparable training settings.
\\
\textbf{Performance with Inception-BN:}
For fair comparison, we also evaluate using Inception-V1 with Batch-Normalization \cite{googlenetv2}. We follow the standard pipeline (see e.g. \cite{proxynca,softriple}), utilizing Adam \cite{adam} with images resized and random cropped to 224x224. The learning rate is set to $10^{-5}$. We retain the size of the policy network and other hyperparameters. The results on CUB200-2011\cite{cub200-2011} and CARS196\cite{cars196} are listed in Table \ref{tab:ibn_sota}. On CUB200, we achieve results competitive to previous state-of-the-art methods. On CARS196, we achieve a significant boost over baseline values and competitive performance to the state-of-the-art.\\
\textbf{Validation set $\mathcal{I}_{\text{val}}$:} 
The validation set $\mathcal{I}_{val}$ is sampled from the training set $\mathcal{I}_{train}$, composed as either a fixed disjoint, held-back subset or repetitively re-sampled from $\mathcal{I}_{train}$ during training. Further, we can sample $\mathcal{I}_{val}$ across all classes or include entire classes. We found (Tab. \ref{tab:validation} (d)) that sampling $\mathcal{I}_{val}$ from each class works much better than doing it per class. Further, resampling $\mathcal{I}_{val}$ provides no significant benefit at the cost of an additional hyperparameter to tune.
\\
\textbf{Composition of states $s$ and target metric $e$:} Choosing meaningful target metrics $e(\phi(\cdot;\zeta), \mathcal{I}_{\text{val}})$ for computing rewards $r$ and a representative composition of the training state $s$ increases the utility of our learned policy $\pi_\theta$. To this end, Tab. \ref{tab:suppl_state_target_ablation} compares different combinations of state compositions and employed target metrics $e$. We observe that incorporating information about the current structure of the embedding space $\Phi$ into $s$, such as intra- and inter-class distances, is most crucial for effective learning and adaptation. Moreover, also incorporating performance metrics into $s$ which directly represent the current performance of the model $\phi$, e.g. Recall@1 or NMI, additional adds some useful information.
\\
\textbf{Frequency of updating $\pi_\theta$:} We compute the reward $r$ for an adjustment $a$ to $p(I_n|I_a)$ every $M$ DML training iterations. High values of $M$ reduce the variance of the rewards $r$, however, at the cost of slow policy updates which result in potentially large discrepancies to updating $\phi$. Tab. \ref{tab:suppl_ablations} (a) shows that choosing $M$ from the range $[30,70]$ results in a good trade-off between the stability of $r$ and the adaptation of $p(I_n|I_a)$ to $\phi$. Moreover, we also show the result for setting $M = \infty$, i.e. using the initial distribution throughout training without adaptation. Fixing this distribution performs worse than the reference method Margin loss with static distance-based sampling\cite{margin}. Nevertheless, frequently adjusting $p(I_n|I_a)$ leads to significant superior performance, which indicates that our policy $\pi_\theta$ effectively adapts $p(I_n|I_a)$ to the training state of $\phi$.
\\
\textbf{Importance of long-term information for states $s$:} For optimal learning, $s$ should not only contain information about the current training state of $\phi$, but also about some history of the learning process. Therefore, we compose $s$ of a set of running averages over different lengths $\mathcal{R}$ for various training state components, as discussed in the implementation details of the main paper. Tab. \ref{tab:suppl_ablations} (b) confirms the importance of long-term information for stable adaptation and learning. Moreover, we see that the set of moving averages $\mathcal{R}=\{2, 8, 16, 32\}$ works best.

\begin{table*}[h]
    \setlength\tabcolsep{1.5pt}
    \centering
     \begin{tabular}{l|c|ccc|c||ccc|c|}
        \toprule
        Dataset & & \multicolumn{4}{c}{CUB200-2011\cite{cub200-2011}} & \multicolumn{4}{c}{CARS196\cite{cars196}}\\ 
        \hline
        Approach & Dim & R@1 & R@2 & R@4  & NMI & R@1 & R@2 & R@4  & NMI\\
        \midrule
        HTG\cite{htg} & 512 & 59.5 & 71.8 & 81.3 & -                                  & 76.5 & 84.7 & 90.4 & - \\
        HDML\cite{hardness-aware} & 512 & 53.7 & 65.7 & 76.7 & 62.6                 & 79.1 & 87.1 & 92.1 & 69.7\\
        HTL\cite{htl} & 512 & 57.1 & 68.8 & 78.7 & -                                & 81.4 & 88.0 & 92.7  & - \\
        DVML\cite{dvml} & 512 & 52.7 & 65.1 & 75.5 & 61.4                           & 82.0 & 88.4 & 93.3 & 67.6\\
        A-BIER\cite{abier} & 512 & 57.5 & 68.7 & 78.3 & -                           & 82.0 & 89.0 & 93.2  & - \\
        MIC\cite{mic} & 128 & 66.1 & 76.8 & 85.6 & 69.7                             & 82.6 & 89.1 & 93.2 & 68.4 \\
        D\&C\cite{Sanakoyeu_2019_CVPR} & 128 & 65.9 & 76.6 & 84.4 & 69.6            & 84.6 & 90.7 & 94.1 & 70.3\\
        Margin\cite{margin} & 128 & 63.6 & 74.4 & 83.1 & 69.0                     & 79.6 & 86.5 & 90.1 & 69.1\\            
        \textbf{Reimpl. Margin\cite{margin}, IBN} & 512 & 63.8 & 75.3 & 84.7 & 67.9     & 79.7 & 86.9 & 91.4 & 67.2 \\        
        \textbf{Ours(Margin\cite{margin} + \textit{PADS}, IBN)} & 512 & 66.6 & 77.2 & 85.6 & 68.5     & 81.7 & 88.3 & 93.0 & 68.2 \\
        \hline
        \multicolumn{10}{l}{Significant increase in network parameter:} \\      
        \hline
        HORDE\cite{horde}+Contr.\cite{contrastive} & 512 & 66.3 & 76.7 & 84.7 & -                   & 83.9 & 90.3 & 94.1 & - \\
        SOFT-TRIPLE\cite{softriple} & 512 & 65.4 & 76.4 & 84.5& -             & 84.5 & 90.7 & 94.5 & 70.1\\
        \hline
        \multicolumn{10}{l}{Ensemble Methods:} \\
        \hline
        Rank\cite{rankedlist} & 1536 & 61.3 & 72.7 & 82.7 & 66.1    & 82.1 & 89.3 & 93.7& 71.8\\
        DREML\cite{dreml} & 9216 & 63.9 & 75.0 & 83.1 & 67.8    & \textbf{86.0} & \textbf{91.7} & \textbf{95.0} & \textbf{76.4}\\
        ABE\cite{abe} & 512 & 60.6 & 71.5 & 79.8 & -        & 85.2 & 90.5 & 94.0 & -\\
        \bottomrule
    \end{tabular}
    \caption{Comparison to the state-of-the-art DML methods on CUB200-2011\cite{cub200-2011} and CARS196\cite{cars196} using the Inception-BN Backbone (see e.g. \cite{proxynca,softriple}) and embedding dimension of 512.}
    \label{tab:ibn_sota}
\end{table*}
\begin{table}[t]
 \centering
 \begin{tabular}{l|c|c|c|c}
    \toprule
    \textbf{Validation Set:} & $\mathcal{I}^{\text{By}}_{\text{val}}$ & $\mathcal{I}^{\text{Per}}_{\text{val}}$ & $\mathcal{I}^{\text{By, R}}_{\text{val}}$ & $\mathcal{I}^{\text{Per, R}}_{\text{val}}$ \\
    \midrule
    Recall@1 & $62.6$ & $65.7$ & $63.0$ & $\mathbf{65.8}$\\
    NMI      & $67.7$ & $69.2$ & $67.8$ & $\mathbf{69.6}$\\    
    \bottomrule
\end{tabular}
\caption{Composition of $\mathcal{I}_{\text{val}}$. Superscript $By$/$Per$ denotes usage of entire classes/sampling across classes. $R$ denotes re-sampling during training with best found frequency of $\frac{1}{\text{50 epochs}}$.}
\label{tab:validation}
\end{table}


\begin{table}
\begin{subtable}{0.45\textwidth}
  \centering
  \begin{tabular}{l|c|c|c}
    \toprule
    $\frac{\text{Reward metrics} \; e}{\text{Composition of state} \; s}$ & NMI & R@1 & R@1 + NMI \\
    \midrule
    \multirow{2}{*}{Recall, Dist., NMI}  & 63.9 & 65.5 & \textbf{65.6} \\
                                         & 68.5 & 68.9 & 69.2 \\    
    \midrule
    \multirow{2}{*}{Recall, Dist.}       & 65.0 & \textbf{65.7} & 64.4 \\
                                         & 68.5 & \textbf{69.2} & \textbf{69.4} \\    
    \midrule
    \multirow{2}{*}{Recall, NMI}         & 63.7 & 63.9 & 64.2 \\
                                         & 68.4 & 68.2 & 68.5 \\    
    \midrule
    \multirow{2}{*}{Dist., NMI}          & 65.3 & 65.3 & 65.1\\
                                         & 68.8 & 68.7 & 68.5\\    
    \midrule
    \multirow{2}{*}{Dist.}               & 65.3 & 65.5 & 64.3 \\
                                         & 68.8 & 69.1 & 68.6 \\    
    \midrule
    \multirow{2}{*}{Recall}              & 64.2 & 65.1 & 64.9 \\
                                         & 67.8 & 69.0 & 68.4 \\
    \midrule
    \multirow{2}{*}{NMI}                 & 64.3 & 64.8 & 63.9 \\
                                         & 68.7 & \textbf{69.2} & 68.4 \\                                                
    \bottomrule
\end{tabular}
\end{subtable}
\caption{Comparison of different compositions of the training state $s$ and reward metric $e$. \textit{Dist.} denotes average intra- and inter-class distances. Recall in state composition denotes all Recall@k-values, whereas for the target metric only Recall@1 was utilized.
}
\label{tab:suppl_state_target_ablation}
\end{table}

\begin{table}
\begin{subtable}{0.45\textwidth}
  \centering
  \begin{tabular}{l|c|c|c|c|c|c|c}
    \toprule
    $M$ & 10 & 30 & 50 & 70 & 100 & $\infty$ & \cite{margin} \\
    \midrule
    R@1  & 64.4 & \textbf{65.7} & 65.4 & 65.2 & 65.1 & 61.9 & 63.5 \\
    NMI     & 68.3 & \textbf{69.2} & 69.2 & 68.9 & 69.0 & 67.0 & 68.1 \\    
    \bottomrule
\end{tabular}
\caption{Evaluation of the policy update frequency $M$.}
\end{subtable}
\begin{subtable}{0.45\textwidth}
  \centering
  \begin{tabular}{l|c|c|c|c}
    \toprule
    $\mathcal{R}$ & 2 & 2, 32 & 2, 8, 16, 32 & 2, 8, 16, 32, 64 \\
    \midrule
    R@1 & 64.5 & 65.4 & \textbf{65.7} & 65.6\\
    NMI & 68.6 & 69.1 & 69.2 & \textbf{69.3}\\
    \bottomrule
  \end{tabular}
\caption{Evaluation of various sets $\mathcal{R}$ of moving average lengths.}
\end{subtable}
\caption{Ablation experiments: (a) evaluates the influence of the number of DML iterations $M$ performed before updating the policy $\pi_\theta$ using a reward $r$ and, thus, the update frequency of $\pi_\theta$. (b) analyzes the benefit of long-term learning progress information added to training states $s$ by means of using various moving average lengths $\mathcal{R}$.
}
\label{tab:suppl_ablations}
\end{table}

\begin{figure}[t]
\begin{center}
\includegraphics[width=0.45\textwidth]{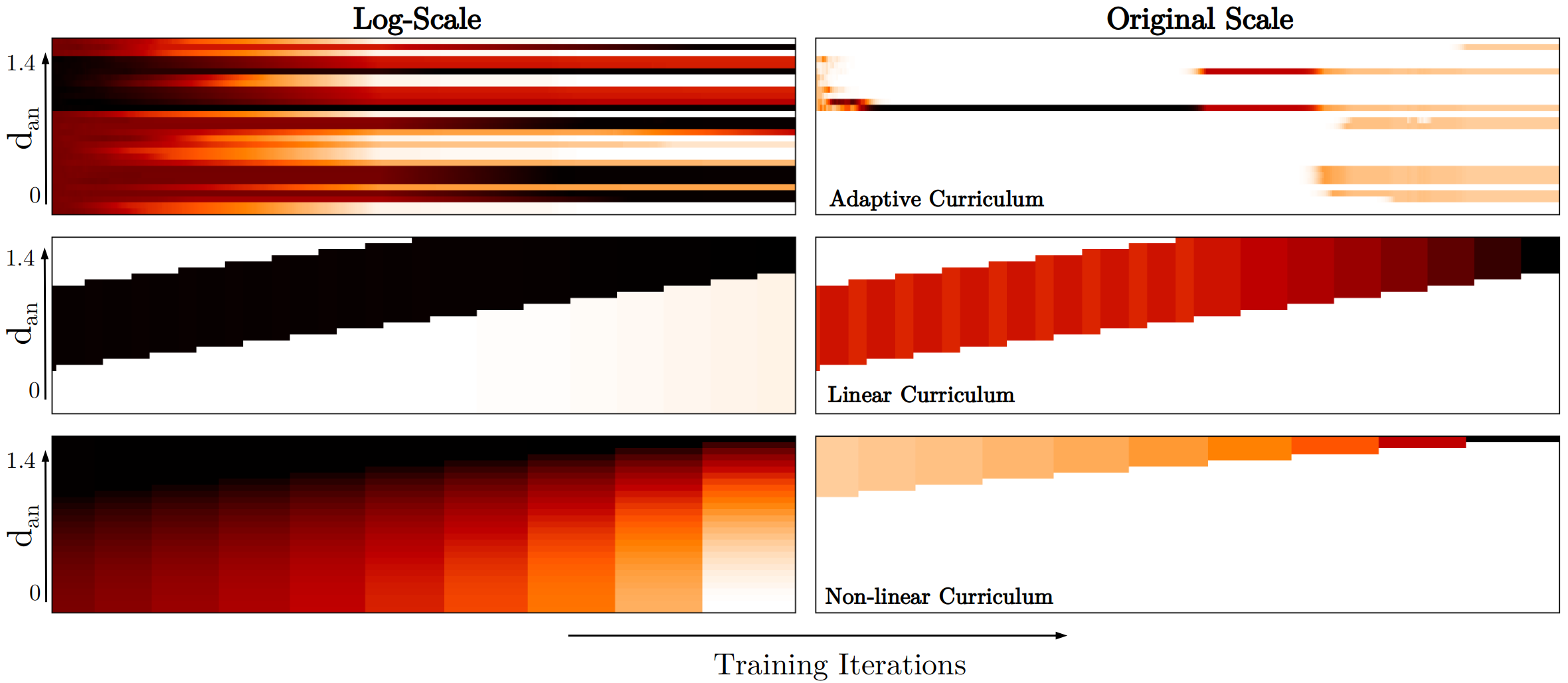}
\end{center}
\caption{\textit{Visual comparison between fixed sampling curriculums and a learned progression of $p(I_n|I_a)$ by PADS.} Left: log-scale over $p(I_n|I_a)$, right: original scale. Top row: learned sampling schedule (PADS); middle row: linear shift of a sampling interval from semihard\cite{semihard} negatives to hard negatives; bottom row: shifting a static distance-based sampling\cite{margin} to gradually sample harder negatives.}
\label{fig:curriculum}
\end{figure}

\section{Curriculum Evaluations}
In Fig. \ref{fig:curriculum} we visually illustrate the fixed curriculum schedules which we applied for the comparison experiment in Sec. 5.3 of our main paper. We evaluated various schedules - Linear progression of sampling intervals starting at semi-hard negatives going to hard negatives, and progressively moving $\mathcal{U}$-dist\cite{margin} towards harder negatives. The schedules visualized were among the best performing ones to work for both CUB200 and CARS196 dataset.


\section{Comparison of RL Algorithms}
We evaluate the applicability of the following RL algorithms for optimizing our policy $\pi_\theta$ (Eq. 4 in the main paper):

\begin{table}[h]
  \setlength\tabcolsep{1.4pt}
  \centering
  \begin{tabular}{l|cc}
    \toprule
    Approach & R@1 & NMI \\
    \midrule
    Margin\cite{margin}   & $63.5$ & $68.1$\\
    \midrule
    REINFORCE               & $64.2$ & $68.5$ \\
    REINFORCE, EMA          & $64.8$ & $68.9$ \\
    REINFORCE, A2C          & $65.0$ & $69.0$ \\    
    PPO, EMA                & $65.4$ & $69.0$ \\
    PPO, A2C                & $\mathbf{65.7}$ & $\mathbf{69.2}$ \\
    Q-Learn                 & $63.2$ & $67.9$ \\
    Q-Learn, PR/2-Step      & $64.9$ & $68.5$ \\
    \bottomrule
  \end{tabular}
\caption{Comparison of different RL algorithms. For policy-based algorithms (REINFORCE, PPO) we either use Exponential Moving Average (EMA) as a variance-reducing baseline or employ Advantage Actor Critic (A2C). In addition, we also evaluate Q-Learning methods (vanilla and Rainbow Q-Learning). For the Rainbow setup we use Priority Replay and 2-Step value approximation. Margin loss\cite{margin} is used as a representative reference for static sampling strategies.
}
\label{tab:ablation_rl_method}
\end{table}

\begin{itemize}
    \item REINFORCE algorithm\cite{reinforce} with and without Exponential Moving Average (EMA)
    \item Advantage Actor Critic (A2C)\cite{Sutton1998}
    \item Rainbow Q-Learning\cite{rainbow_dqn} without extensions (vanilla) and using Priority Replay and 2-Step updates
    \item Proximal Policy Optimization (PPO)\cite{ppo} applied to REINFORCE with EMA and to A2C.
\end{itemize}

\noindent
For a comparable evaluation setting we use the CUB200-2011\cite{cub200-2011} dataset without learning rate scheduling and fixed 150 epochs of training. Within this setup, the hyperparameters related to each method are optimized via cross-validation. Tab. \ref{tab:ablation_rl_method} shows that 
all methods, except for vanilla Q-Learning, result in an adjustment policy $\pi_\theta$ for $p(I_n|I_a)$ which outperforms static sampling strategies. Moreover, policy-based methods in general perform better than Q-Learning based methods with PPO being the best performing algorithm. We attribute this to the reduced search space (Q-Learning methods need to evaluate in state-actions space, unlike policy-methods, which work directly over the action space), as well as not employing replay buffers, i.e. not acting off-policy, since state-action pairs of previous training iterations may no longer be representative for current training stages.

\section{Qualitative UMAP Visualization}
\begin{figure}
\begin{center}
\includegraphics[width=0.49\textwidth]{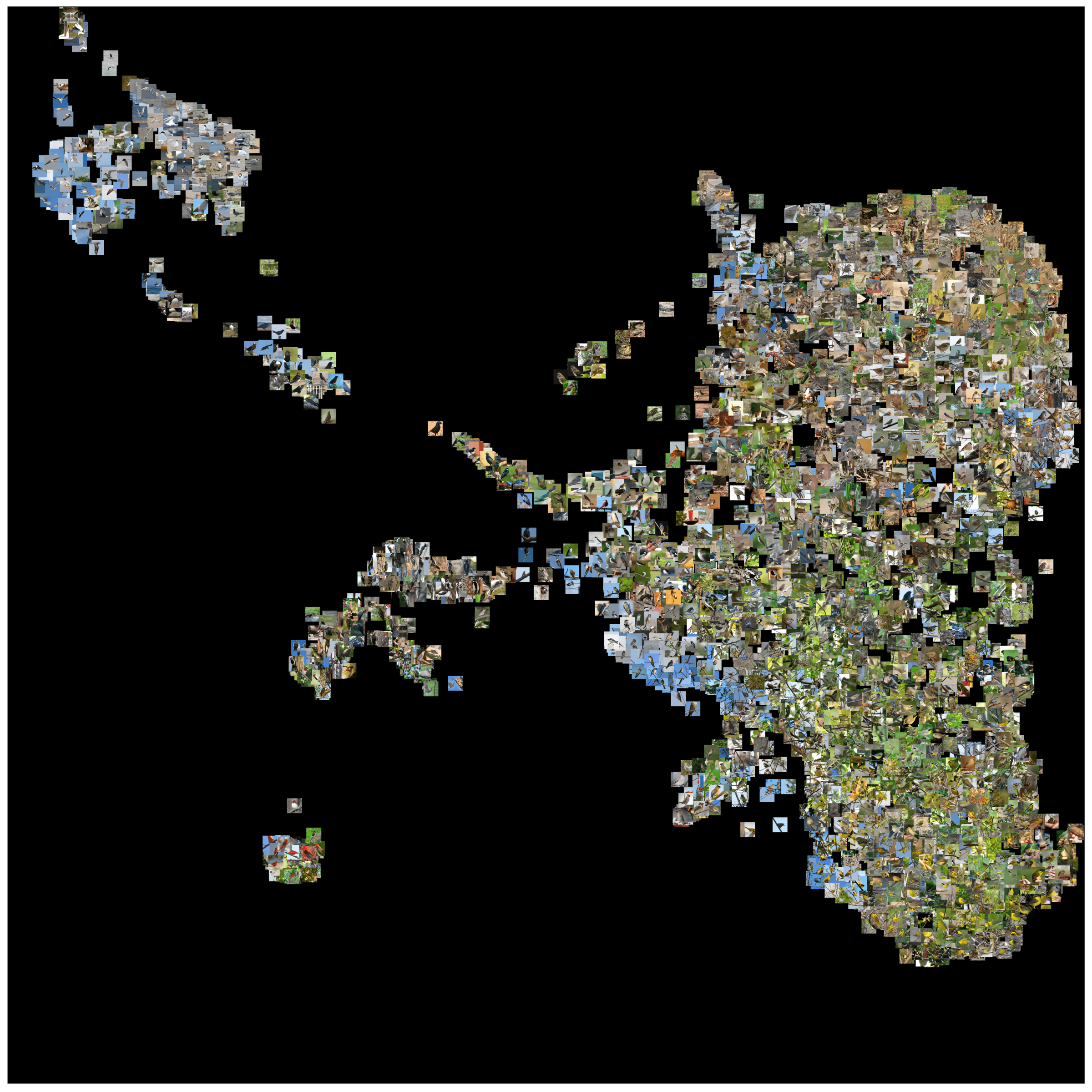}
\end{center}
  \caption{\textit{UMAP embedding} based on the image embeddings $\phi(\cdot;\zeta)$ obtained from our proposed approach on CUB200-2011\cite{cub200-2011} (Test Set).}
\label{fig:umap}
\end{figure}

Figure \ref{fig:umap} shows a UMAP\cite{umap} embedding of test image features for CUB200-2011\cite{cub200-2011} learned by our model using PADS. We can see clear groupings for birds of the same and similar classes. Clusterings based on similar background is primarily due to dataset bias, e.g. certain types of birds occur only in conjunction with specific backgrounds.

\section{Pseudo-Code}
Algorithm \ref{alg:algorithm_short} gives an overview of our proposed \textit{PADS} approach using PPO with A2C as underlying RL method.\\
Before training, our sampling distributions $p(I_n|I_a)$ is initialized with an initial distribution. Further, we initialize both the adjustment policy $\pi_\theta$ and the pre-update auxiliary policy $\pi_\theta^{old}$ for estimating the PPO probability ratio. Then, DML training is performed using triplets with random anchor-positive pairs and sampled negatives from the current sampling distribution $p(I_n|I_a)$. After $M$ iterations, all reward and state metrics $\mathcal{E}, \mathcal{E}^*$ are computed on the embeddings $\phi(\cdot;\zeta)$ of $\mathcal{I}_{val}$. These values are aggregated in a training reward $r$ and input state $s$. While $r$ is used to update the current policy $\pi_\theta$, $s$ is fed into the updated policy to estimate adjustments $a$ to the sampling distribution $p(I_n|I_a)$. Finally, after $M^{\text{old}}$ iterations (e.g. we set to $M^{\text{old}}=3$) $\pi_\theta^{old}$ is updated with the current policy weights $\theta$.

\begin{algorithm}
\caption{Training one epoch via \textit{PADS} by PPO}

\SetKwFunction{UpdatePolicy}{UpdatePolicy}
\SetKwFunction{TrainDML}{TrainDML}
\SetKwFunction{Sample}{Sample}
\SetKwFunction{PPOLoss}{PPOLoss}
\SetKwFunction{Split}{Split}
\SetKwFunction{Backward}{Backward}
\SetKwFunction{SampleBatch}{SampleBatch}
\SetKwFunction{SampleActions}{SampleActions}
\SetKwFunction{InitPolicy}{InitPolicy}
\SetKwFunction{GetState}{GetState}
\SetKwFunction{ValMetrics}{ValMetrics}
\SetKwFunction{GetReward}{GetReward}
\SetKwFunction{Copy}{Copy}
\SetKwFunction{Adjust}{Adjust}
\SetKwInOut{Input}{input}
\SetKwInOut{Init}{initialization}
\SetKwInOut{Constants}{parameters}
\SetKwRepeat{Repeat}{repeat}{until}
\SetKwInOut{Input}{Input}
\SetKwInOut{Parameter}{Parameter}

\SetAlgoLined
\Input{$\mathcal{I}_{\text{train}}$, $\mathcal{I}_{\text{val}}$, Train labels $\mathcal{Y}_{\text{train}}$, Val. labels $\mathcal{Y}_{\text{val}}$, total iterations $n_e$}

\Parameter{Reward metrics $\mathcal{E}$, State metrics $\mathcal{E^*}$ + running average lengths $\mathcal{R}$, Num. of bins $K$, multiplier $\{\alpha, \beta\}$, $p_{\text{init}}(I_n|I_a)$, num. of iterations before updates $M$, $M^{\text{old}}$}  

\text{ }
\newline

\tcp{\textbf{Initialization}}
$p(I_n|I_a) \leftarrow p_{\text{init}}(I_n|I_a)$  

$\pi_\theta \leftarrow \InitPolicy(K, \alpha, \beta)$  

$\pi_\theta^{old} \leftarrow \Copy(\pi_\theta)$
\newline

\For{i in $n_e/M$}{
    
    \text{ }
    \newline
    
    \tcp{\textbf{Update DML Model}}
    \For{j in $M$}{
        \tcp{within batch $\mathcal{B} \in \mathcal{I}_{\text{train}}$}
        $I_a, I_p \leftarrow \mathcal{B}$
        
        $I_n \sim p(I_n|I_a)$
        
        $\zeta \leftarrow \TrainDML(\{I_a, I_p, I_n\}, \phi(.;\zeta))$
        }
        
    \text{ }
    \newline
    
    \tcp{\textbf{Update policy} $\pi_\theta$}
    $E_i \leftarrow \mathcal{E}(\mathcal{I}_{\text{val}}, \mathcal{Y}_{\text{val}}, \phi(.;\zeta))$

    $E^*_i \leftarrow \mathcal{E}^*(\mathcal{I}_{\text{val}}, \mathcal{Y}_{\text{val}}, \phi(.;\zeta))$
    
    $s_i \leftarrow \GetState(E^*_i, \mathcal{R}, p(I_n|I_a))$
    
    $r \leftarrow \GetReward(E_i, E_{i-1})$
    
    $l_\pi \leftarrow \PPOLoss(\pi_\theta, \pi_\theta^{old}, s_{i-1}, a_{i-1})$
    
    $\theta \leftarrow \UpdatePolicy(l_\pi, \pi_\theta)$
    
    $a_i \sim \pi_\theta(a_i|s_i)$
    
    $p(I_n|I_a) \leftarrow \Adjust(p(I_n|I_a), a_i)$
    \newline
    
    \If{i mod $M^{\text{old}}$ == 0}{
        $\pi_\theta^{old} \leftarrow \Copy(\pi)$
    }
}
    
\label{alg:algorithm_short}
\end{algorithm}

\section{Typical image retrieval failure cases}
\begin{figure}[h]
\begin{center}
\includegraphics[width=0.45\textwidth]{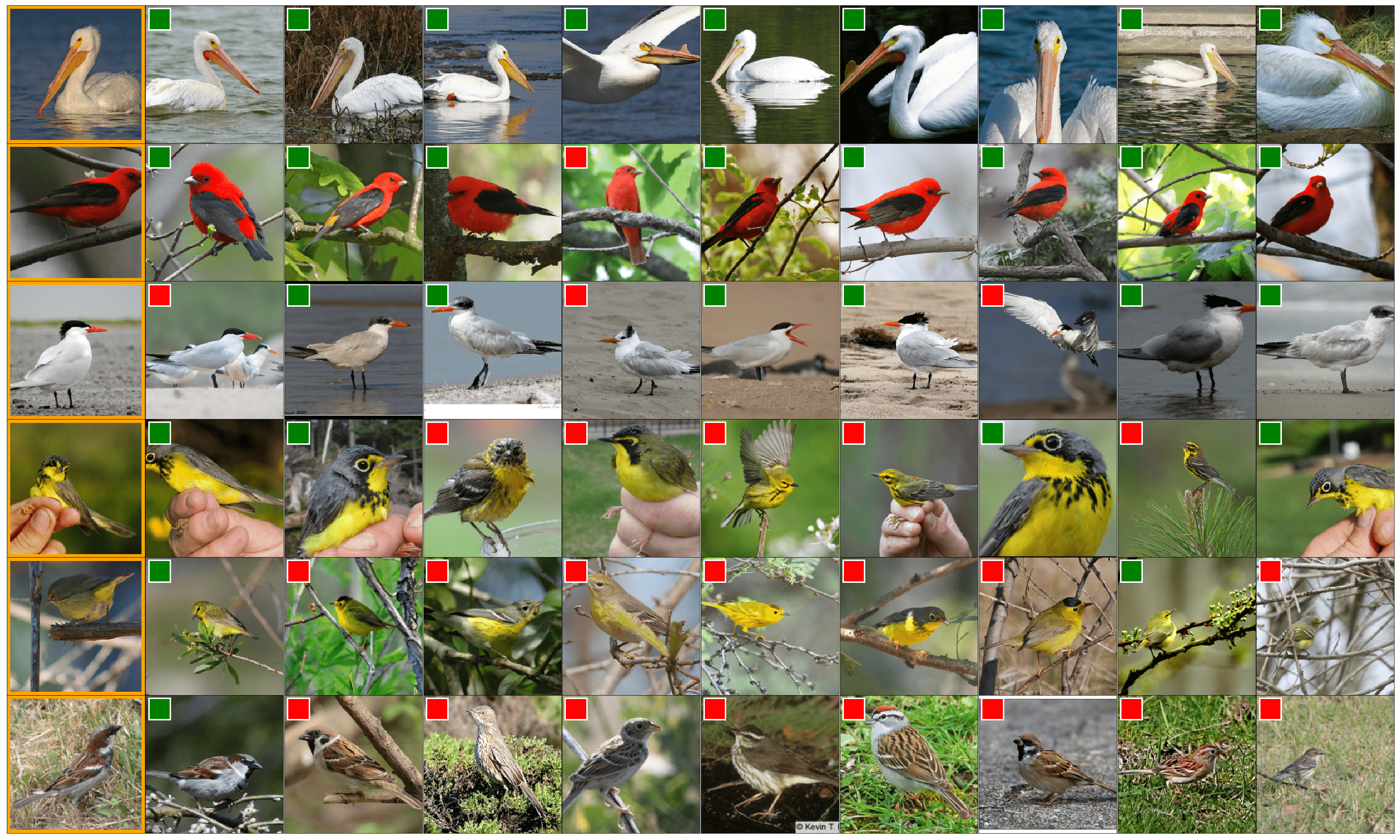}
\end{center}
   \caption{\textit{Selection of good and bad nearest neighbour retrieval cases on CUB200-2011 (Test)}. Orange bounding box marks query images, green/red boxes denote correct/incorrect retrievals.} 
\label{fig:nns}
\end{figure}

Fig. \ref{fig:nns} shows nearest neighbours for good/bad test set retrievals. Even though the nearest neighbors do not always share the same class label as the anchor, all neighbors are very similar to the bird species depicted in the anchor images. Failures are due to very subtle differences.

\end{document}